\documentclass[letterpaper, 10 pt, conference]{ieeeconf}  % Comment this line out if you need a4paper

\usepackage{tabularx}
\usepackage{hyperref}
\usepackage{booktabs}

\IEEEoverridecommandlockouts                              % This command is only needed if 
\usepackage{graphicx} % for pdf, bitmapped graphics files
\usepackage{amsmath} % assumes amsmath package installed
\usepackage{amssymb}  % assumes amsmath package installed

\usepackage{multirow} %todo: check if allowed to add
\usepackage{graphicx}
\usepackage{subcaption} % Essential for (a) and (b) sub-figures
\usepackage{tikz}
\usetikzlibrary{arrows.meta, shapes.geometric, decorations.pathreplacing, patterns, calc, positioning}
\usepackage{algorithm}
\usepackage{algpseudocode}
\usepackage{bm}        % Fixes the undefined \bm error
\usepackage{xurl}

\title{\LARGE \bf
Scalable Production Scheduling: Linear Complexity via Unified Homogeneous Graphs
}

\author{Jonathan Hoss$^{1}$, Moritz Link$^{1}$ and Noah Klarmann$^{1}$%  
\thanks{*The work has been performed in the Cynergy4MIE project (GA. 101140226)}%  
\thanks{$^{1}$Faculty of Management and Engineering, Rosenheim Technical University of Applied Sciences, Germany, Correspondence: Jonathan Hoss, {\tt\small jonathan.hoss@th-rosenheim.de}}%  
}

\begin{document}
\maketitle
\thispagestyle{empty}
\pagestyle{empty}

\begin{abstract}
Efficiently solving the Job Shop Scheduling Problem in real-world industrial applications requires policies that are both computationally lean and topologically robust. While Reinforcement Learning has shown potential in automating dispatching rules, existing models often struggle with a scalability bottleneck caused by quadratic graph complexity or the architectural overhead of heterogeneous layers. We introduce a unified graph framework that employs feature-based homogenization to project distinct node roles into a shared latent space. This allows a standard homogeneous Graph Isomorphism Network to capture complex resource contention with linear complexity, ensuring low-latency inference for large-scale industrial applications.
Our empirical results demonstrate that our framework achieves state-of-the-art performance while exhibiting consistent zero-shot generalization. We identify the \mbox{job-to-machine} ratio as the primary driver of policy effectiveness, rather than absolute problem size. Based on this, we propose a hypothesis of structural saturation, demonstrating that policies trained on critically congested instances ($\mathcal{J} \approx \mathcal{M}$) learn scale-invariant resolution strategies. Agents trained at this saturation point internalize invariant conflict-resolution logic, allowing them to treat massive rectangular instances as a sequential concatenation of saturated sub-problems. This approach eliminates the need for expensive scale-specific retraining and prevents overfitting to statistical shortcuts, providing a robust and efficient pathway for deploying RL solutions in dynamic production environments.

\end{abstract}

\section{Introduction}
The Job Shop Scheduling Problem (JSSP) remains a cornerstone challenge in operations research and combinatorial optimization, representing the optimal allocation of shared resources within strictly constrained environments. In industrial execution, scheduling is not a static combinatorial optimization problem, but a dynamic control problem characterized by strict causality: decisions taken at time $t$ are irrevocable and must adapt continuously to evolving shop floor conditions. 
While exact optimization methods guarantee optimal schedules, their computational complexity grows exponentially with problem size, rendering them unsuitable for online application in large-scale production environments. Therefore, industrial practice is currently dominated by Priority Dispatching Rules (PDRs), which satisfy stringent latency constraints. However, these heuristics are inherently myopic and static, limiting their ability to react to dynamic bottlenecks and complex downstream interactions.

Reinforcement Learning (RL) provides a means to bridge the gap between simple, fast heuristics and the need for adaptive, foresighted scheduling policies by learning directly from interaction with the evolving shop floor state. However, a central challenge in the industrial application of RL lies in the ability to deploy learned policies beyond their training distribution, as production systems vary widely in size, topology, and resource configuration. 

Drawing on the principle of making "infinite use of finite means", Battaglia et al. \cite{battagliaRelationalInductiveBiases2018} argue that achieving such human-like combinatorial generalization requires architectures with strong relational inductive biases. These biases enable the systematic reuse of learned computations across varying structures, rather than memorization of specific configurations. Graph Neural Networks (GNNs) naturally provide this property, as their functions are shared and reused across all nodes and edges, allowing policies to generalize systematically to unseen topologies. 

The JSSP can be fully represented using the disjunctive graph formulation, which models machine contention by connecting every operation to all other operations on the same machine. While the JSSP becomes solvable within this representation by directing these machine constraint edges, this formulation creates dense cliques with quadratic edge complexity. As problem size increases, the resulting message-passing cost grows rapidly, creating a severe computational bottleneck that hinders scalability and practical application.

Recent strategies have adopted heterogeneous graph representations to explicitly model resource constraints, successfully reducing structural density. However, this shifts the complexity from the data to the architecture: these models require separate parameters for different entity types. This architectural overhead increases memory footprint and creates inference latency, effectively replacing a graph scalability bottleneck with a model efficiency bottleneck.

In this work, we propose a unified graph framework that retains the benefits of a heterogeneous graph representation while eliminating the computational overhead of heterogeneous layers. We model the shop floor as a sparse heterogeneous graph to efficiently capture resource contention, but employ a strategy of feature-based homogenization. By projecting distinct node roles into a unified feature space, we enable a standard homogeneous GNN to process the structurally heterogeneous graph. This effectively resolves the trade-off between representational fidelity and computational efficiency and enables scalable policy learning with strong relational inductive biases.

Our core contributions are as follows: 
\begin{itemize}
\item \textbf{Linear-Complexity Unified Graph:} We introduce a unified graph representation that models machines as first-class entities and captures machine–operation interactions via a sparse bipartite structure, reducing dense disjunctive cliques to linear complexity and enabling scalable online inference for industrial application.

\item \textbf{Feature-Based Homogenization:} We propose a feature-based homogenization strategy that embeds heterogeneous node roles into a shared latent space, allowing the use of a standard homogeneous Graph Isomorphism Network (GIN) without heterogeneous layers or type-specific parameters.

\item \textbf{Structural Saturation Hypothesis:} We identify a critical problem scale at which congestion patterns in JSSP instances become fully developed and structurally stable and show that training at this scale induces \mbox{scale-invariant} scheduling behavior.

\item \textbf{Industrial-Scale Zero-Shot Generalization:} We demonstrate that a single policy trained at the structural saturation point generalizes in a zero-shot manner across problem sizes, eliminating the need for size-specific retraining and satisfying key industrial constraints on scalability and deployment.
\end{itemize}

\section{Related Work}
\subsection{Reinforcement Learning for Scheduling}
In the context of JSSP, Reinforcement Learning typically formulates scheduling as a constructive process. Starting from an empty schedule, an agent sequentially selects operations to assign to machines, allowing the policy to adapt to the dynamic state $s_t$ without requiring a full recalculation of the schedule. A primary distinction in the literature lies in how this system state is represented to the learning agent, broadly dividing approaches into fixed-dimensional and graph-based architectures.

\textbf{Fixed-Dimensional Representations:} 
A foundational line of research formulates the scheduling state using Euclidean architectures, differing primarily in how domain knowledge is encoded. Tassel et al. \cite{tasselReinforcementLearningEnvironment2021} relied on explicit feature engineering, utilizing domain expertise to manually calculate high-level attributes for processing by Multi-Layer Perceptrons (MLPs). In contrast, Qiao et al. \cite{qiaoOptimizationJobShop2024} focused on structural representation, organizing JSSP problem instances into multi-channel image-like tensors. This formulation allows Convolutional Neural Networks (CNNs) to implicitly extract relational patterns via spatial filters. Despite these methodological differences, both streams face a representation bottleneck: they rely on input representations with fixed dimensionality, requiring zero-padding or retraining to handle varying instance sizes. This structural rigidity effectively precludes zero-shot generalization to unseen problem scales.

\textbf{Graph Representations:} To enable size-agnostic reasoning, state-of-the-art methods have adopted GNNs. These architectures utilize topological state representations that allow policies trained on small instances to transfer to larger problems. 
Zhang et al. \cite{zhangLearningDispatchJob2020} proposed the "Learning to Dispatch" framework, which reduces the computational burden of full disjunctive cliques by using a dynamic graph representation. In this formulation, machine edges connect only operations currently scheduled or immediately eligible. While this limits complexity, it also restricts explicit reasoning about the waiting queue topology, as future contention is not structurally encoded. Moreover, the state transitions allow retroactive operation insertions, enabling previously made dispatching decisions to be altered. Although sufficient for offline schedule generation, this property precludes application in online or dynamic settings where past decisions must remain fixed.

In contrast, Park et al. \cite{parkLearningScheduleJobshop2021} retained the complete disjunctive graph representation to capture full global dependencies. While effective on smaller benchmarks, they explicitly acknowledge that the resulting quadratic edge complexity causes computational costs to rise rapidly with the number of operations, limiting scalability for large-scale instances.

\textbf{Explicit Resource Modeling and Heterogeneity:}
To address the topological inefficiency of disjunctive graphs, recent research has shifted toward explicit resource modeling, where machines are instantiated as distinct nodes within the graph. This evolution has been largely driven by the routing complexities of the Flexible JSSP. Approaches such as those by Song et al. \cite{songFlexibleJobShopScheduling2023} and Wang et al. \cite{wangFlexibleJobShop2024} utilize heterogeneous GNNs to capture operation-machine interactions, relying on disjoint parameter sets to process the distinct node types. Similarly, Hameed and Schwung \cite{hameedGraphNeuralNetworksbased2023} model the shop floor as a bipartite graph. Their framework assigns a learning agent to each individual machine. This introduces multi-agent coordination overhead as it decomposes the problem into decentralized local optimizations.

Critically, these methods often combine structural heterogeneity with architectural heterogeneity. Using specialized heterogeneous GNNs necessitates separate weight matrices for each edge type, substantially increasing learnable parameters and computational overhead. Our work seeks to decouple this relationship: we leverage explicit resource nodes for their sparse linear topology, but process the graph with a unified, homogeneous backbone. Aligning with findings by Wang et al. \cite{wangEnablingHomogeneousGNNs2023a} on the expressivity of homogeneous GNNs, we aim to achieve superior scalability without the computational burden of heterogeneous architectures.

\section{Preliminaries}
\subsection{Job Shop Scheduling Problem}
The JSSP involves scheduling a set of jobs $\mathcal{J}$ on a set of machines $\mathcal{M}$. 
Each job $J_j$ consists of a sequence of operations $(O_{j,1}, \dots, O_{j,M})$, where each operation $O_{j,k}$ is processed on a predefined machine $m_{j,k}$ for a specific duration $p_{j,k}$. A valid schedule must satisfy two conditions: (1) precedence constraints, requiring $S_{j, k+1} \ge S_{j,k} + p_{j,k}$, where $S_{j,k}$ is the start time; and (2) resource constraints, ensuring that for any two operations assigned to the same machine, their processing intervals are disjoint.

The objective is to minimize the makespan $C_{max}$, defined as the completion time of the last operation across all jobs:
\begin{equation}
C_{max} = \max_{j \in \mathcal{J}} (S_{j,M} + p_{j,M})
\end{equation}

\subsection{Reinforcement Learning}
RL is a computational approach to learning through interaction to achieve a goal. As defined by Sutton and Barto \cite{suttonReinforcementLearningIntroduction2018}, the problem is modeled as an agent interacting with an environment over discrete time steps $t$. At each step, the agent observes a state $s_t \in \mathcal{S}$, selects an action $a_t \in \mathcal{A}(s_t)$, and receives a reward $r_{t+1} \in \mathbb{R}$ and the next state $s_{t+1}$. The objective is to learn a policy $\pi(a|s)$ that maximizes the expected return \mbox{$G_t = \sum_{k=0}^{\infty} \gamma^k r_{t+k+1}$}, the discounted sum of future rewards, where $\gamma \in [0, 1]$ is the discount factor.

\section{Method}
This section details our framework for scalable JSSP scheduling. We first formalize the environment as a Markov Decision Process (MDP), followed by a heterogeneous graph representation to ensure linear topological complexity. We introduce feature-based homogenization to map diverse entities into a unified space, enabling a computationally efficient neural architecture based on a homogeneous GIN. Finally, we implement an actor-critic framework trained via Proximal Policy Optimization (PPO) \cite{schulman2017proximal} to derive adaptive, scale-invariant scheduling policies.

\subsection{Markov Decision Process Formulation}
\label{sec:mdp}

We formalize the constructive scheduling process as a finite-horizon MDP defined by the tuple $(\mathcal{S}, \mathcal{A}, \mathcal{P}, r, \gamma)$, representing the state space, action space, transition dynamics, reward function, and discount factor, respectively.

The state space $\mathcal{S}$ is represented by a sequence of directed heterogeneous graphs $\{G_t\}$ that capture the spatial and temporal configuration of the shop floor. 

At each step, the agent selects an action $a_t$ from the action space $\mathcal{A}_t = \{ O_{j,k} \mid O_{j,k-1} \in \mathcal{C} \land O_{j,k} \notin \mathcal{C} \}$, where $\mathcal{C}$ is the set of completed operations, ensuring that an operation only becomes eligible once its direct predecessor is finished. We employ action masking to restrict the policy’s output distribution to this dynamic set. 

The transition dynamics $\mathcal{P}$ are deterministic, where selecting an operation $O_{j,k}$ sets its start time according to $S_{j,k} = \max(C_{j,k-1}, \text{Avail}(\mathcal{M}_k))$. This assignment updates the machine's availability and evolves the graph topology to activate the successor operation $O_{j,k+1}$.

The reward function $r$ combines a sparse terminal signal with dense shaping. At the terminal state ($t=T$), the agent receives a sparse terminal reward based on the proximity of the achieved makespan to the theoretical lower bound ($LB$): 

\begin{equation}
    r_T =  \lambda_{\text{term}} \cdot \frac{LB}{C_{\max}},
\end{equation}
where $\lambda_{\text{term}}$ is a scaling factor. This term serves as the ground-truth objective, directly aligning the policy with global makespan minimization.

\begin{figure}
\vspace*{0.1cm}
    \centering
    \begin{tikzpicture}[font=\footnotesize, >=stealth, xscale=0.9, yscale=0.9]
        
        % Global Styles
        \tikzset{
            op_bar/.style={draw=black, thick, fill=gray!20},
            measure_line/.style={densely dashed, thin, draw=black!70},
            axis_line_arr/.style={->, thick},
            axis_line/.style={-, thick},
            axis_label/.style={font=\scriptsize} % Style for the new labels
        }

        % ==========================
        % LEFT: HIGH SPREAD
        % ==========================
        \begin{scope}
            % Title lowered from 2.6 to 1.7
            \node[] at (2.0, 1.7) {High Spread};
            
            % Axes (Y-axis reduced from 2.5 to 1.5)
            \draw[axis_line_arr] (0,0) -- (4.2,0) node[below left, axis_label] {Time};
            \draw[axis_line] (0,0) -- (0,1.5) node[midway, above, rotate=90, axis_label] {Machines};

            % Operations (Height reduced to 0.3, positions compacted)
            % Old Top: (0,1.6) rectangle (3.8,2.2) -> New: (0,1.0) rectangle (3.8,1.3)
            \draw[op_bar] (0,1.0) rectangle (3.4,1.3); 
            % Old Bottom: (0,0.6) rectangle (1.0,1.2) -> New: (0,0.4) rectangle (1.0,0.7)
            \draw[op_bar] (0,0.4) rectangle (1.0,0.7); 

            % -- Drop Lines -- (Start point lowered to 1.3)
            \draw[measure_line] (2.0,1.3) -- (2.0,-0.6);
            \draw[measure_line] (3.4,1.3) -- (3.4,-0.6);

            % -- Labels --
            \node[anchor=north] at (1.9, -0.8) {Mean CLB};
            \node[anchor=north] at (3.4, -0.8) {Max CLB};

            % -- Spread Arrow --
            \draw[<->, thick] (2.0,-0.5) -- (3.4,-0.5);
            \node[fill=white, inner sep=1pt, font=\scriptsize] at (2.7, -0.5) {Spread};
        \end{scope}

        % ==========================
        % RIGHT: LOW SPREAD
        % ==========================
        \begin{scope}[xshift=4.6cm] 
            % Title lowered
            \node[] at (2.0, 1.7) {Low Spread};

            % Axes (Y-axis reduced)
            \draw[axis_line_arr] (0,0) -- (4.2,0) node[below left, axis_label] {Time};
            \draw[axis_line] (0,0) -- (0,1.5) node[midway, above, rotate=90, axis_label] {Machines};

            % Operations (Height reduced to 0.3, positions compacted)
            \draw[op_bar] (0,1.0) rectangle (3.4,1.3); 
            \draw[op_bar] (0,0.4) rectangle (2.8,0.7); 

            % -- Drop Lines & Labels --
            
            % 1. Mean (3.1)
            \draw[measure_line] (3.1,1.3) -- (3.1,-0.6);
            \node[anchor=north east, inner sep=2pt] at (3.1, -0.9) {Mean CLB};

            % 2. Max (3.4)
            \draw[measure_line] (3.4,1.3) -- (3.4,-0.6);
            \node[anchor=north west, inner sep=2pt] at (3.1, -0.9) {Max CLB};

            % -- Spread Arrow --
            \draw[<->, thick] (3.1,-0.4) -- (3.4,-0.4);
            \node[left, font=\scriptsize] at (2.9, -0.5) {Spread};
        \end{scope}

    \end{tikzpicture}

    \caption{Structural characteristics of schedule spread. High spread reflects unbalanced job progress, low spread denotes synchronized execution.}
    \label{fig:reward_spread}
    \vspace{-0.4cm} % <--- Adjust this value to pull the text up
\end{figure}
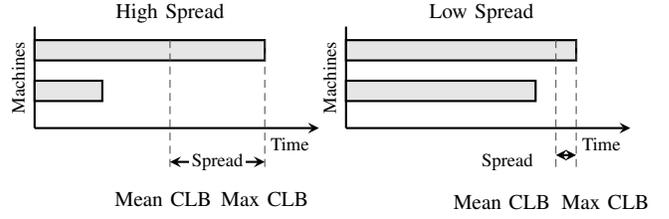

The dense shaping reward is introduced to provide intermediate feedback during schedule construction and to mitigate credit assignment challenges. This component is derived from the spread of operation-level completion lower bounds ($CLB$), where $CLB(o, s_t)$ denotes the theoretical earliest completion time of operation $o$, assuming no future resource contention. We define the spread as the imbalance between the maximum $CLB$ (corresponding to the operation that would finish last) and the mean $CLB$ over all operations in the schedule:

\begin{equation}
    \text{Spread}(s_t)
    =
    \max_{o \in \mathcal{V}_{ops}} CLB(o, s_t)
    -
    \frac{1}{|\mathcal{V}_{ops}|}
    \sum_{o \in \mathcal{V}_{ops}} CLB(o, s_t).
\end{equation}
As illustrated in Figure ~\ref{fig:reward_spread}, a lower spread indicates a more synchronized progression of jobs, whereas a high spread signals that specific jobs are lagging behind, potentially creating future bottlenecks. The stepwise shaping reward is subsequently defined as the reduction in spread between consecutive states, 
\begin{equation}
    \label{eq:shaping_reward}
    r_t
    =
    \lambda_{\text{shape}}
    \big(
    \text{Spread}(s_t) - \text{Spread}(s_{t+1})
    \big).
\end{equation}
By rewarding the reduction of this metric, the agent is incentivized to maintain a balanced schedule, providing a local heuristic that accelerates convergence toward the global objective of makespan minimization.

The theoretical foundation of our dense learning signal is grounded in potential-based reward shaping proposed by Ng et al. \cite{ngPolicyInvarianceReward1999}, ensuring that the additional feedback does not introduce unintended bias into the policy. Under this framework, a shaping reward $F$ preserves the optimal policy of the original MDP if it can be expressed as a difference of potentials $F(s, a, s') = \gamma \Phi(s') - \Phi(s)$.

By defining a potential function over the state space as $\Phi(s) = - \lambda_{\text{shape}} \cdot \text{Spread}(s)$ and assuming a discount factor $\gamma \approx 1$ for the episodic construction, we ensure that our dense reward takes the form of a potential difference:
\begin{equation}
    r_t = \Phi(s_{t+1}) - \Phi(s_t).
\end{equation}
This formulation constitutes a necessary and sufficient condition for policy invariance, guaranteeing that any policy maximizing our shaped reward function remains optimal for the original sparse objective. Consequently, $r_t$ accelerates learning without introducing suboptimal cycles or altering the goal of makespan minimization. 

% TODO added explicit ablation -> jana done
Empirically, we find this reward shaping approach to be decisive for scalability. As a component ablation, we trained agents under the same setup using only the sparse terminal reward. While sparse rewards suffice for smaller problems, agents trained on instances of size $20 \times 20$ and larger failed to converge reliably without the dense signal. This emphasizes that reward shaping is not merely an optimization convenience, but an essential credit-assignment mechanism for larger state spaces.

\subsection{Heterogeneous Graph State Representation}
\label{sec:graph_representation}
We formulate the state $s_t$ as a heterogeneous graph \mbox{$G_t = (\mathcal{V}, \mathcal{E})$}, where the node set $\mathcal{V} = \mathcal{V}_{ops} \cup \mathcal{V}_{mch}$ is divided into operation nodes and machine nodes.

\begin{figure}[tbp]
\vspace*{0.15cm}  % <--- Forces the entire table down
\begin{tikzpicture}[
    % --- Global TikZ Styles ---
    % CHANGED: Reduced horizontal distance from 1.2cm to 0.7cm for tighter packing
    node distance=0.6cm and 0.3cm,
    compact_op/.style={ 
        circle, draw=black, thick, fill=white,
        minimum size=0.55cm, 
        inner sep=0pt, font=\scriptsize\bfseries 
    },
    compact_source/.style={ 
        rectangle, draw=black, thick, minimum size=0.5cm, 
        rounded corners=2pt, fill=gray!15, font=\scriptsize\bfseries
    },
    compact_machine/.style={ 
        rectangle, draw=black, thick, fill=gray!30, 
        minimum size=0.5cm, 
        font=\scriptsize\bfseries
    },
    conj/.style={ 
        ->, >={Stealth[length=1.5mm]}, thick, black,
        shorten >=1pt, shorten <=1pt 
    },
    disj/.style={ 
        <->, >={Stealth[length=1.5mm]}, dashed, thick, black,
        shorten >=1pt, shorten <=1pt 
    }
]

    % ===========================
    % === GRAPH A (Left Side) ===
    % ===========================
    
    % 1. Top Row
    \node[compact_op] (A-J1O1) {$O_{1,1}$};
    \node[compact_op, right=of A-J1O1] (A-J1O2) {$O_{1,2}$};
    \node[compact_op, right=of A-J1O2] (A-J1O3) {$O_{1,3}$};
    
    % 2. Bottom Row
    \node[compact_op, below=of A-J1O1] (A-J2O1) {$O_{2,1}$};
    \node[compact_op, right=of A-J2O1] (A-J2O2) {$O_{2,2}$};
    \node[compact_op, right=of A-J2O2] (A-J2O3) {$O_{2,3}$};
    
    % 3. Source/Sink
    \node[compact_source, left=0.1cm of $(A-J1O1.west)!0.5!(A-J2O1.west)$] (A-S) {S};
    \node[compact_source, right=0.1cm of $(A-J1O3.east)!0.5!(A-J2O3.east)$] (A-T) {E};
    
    % 4. Edges
    \draw[conj] (A-S) -- (A-J1O1.west); \draw[conj] (A-S) -- (A-J2O1.west);
    \draw[conj] (A-J1O1) -- (A-J1O2); \draw[conj] (A-J1O2) -- (A-J1O3);
    \draw[conj] (A-J2O1) -- (A-J2O2); \draw[conj] (A-J2O2) -- (A-J2O3);
    \draw[conj] (A-J1O3.east) -- (A-T); \draw[conj] (A-J2O3.east) -- (A-T);
    
    \draw[disj] (A-J1O1) edge node[pos=0.4, left, xshift=-1pt, font=\tiny] {M1} (A-J2O2);
    \draw[disj] (A-J1O2) edge node[pos=0.4, right, xshift=1pt, font=\tiny] {M2} (A-J2O1);
    \draw[disj] (A-J1O3) -- node[left, font=\tiny] {M3} (A-J2O3);

    % ============================
    % === GRAPH B (Right Side) ===
    % ============================
    
    % 1. Calculate Vertical Alignment
    \path (A-J1O2) -- (A-J2O2) coordinate[midway] (A-Center);

    % 2. Place Center Node of Graph B (M2)
    \node[compact_machine, right=4cm of A-Center] (B-M2) {$M_2$};

    % 3. Build Graph B outwards from M2
    \node[compact_machine, left=0.4cm of B-M2] (B-M1) {$M_1$};
    \node[compact_machine, right=0.4cm of B-M2] (B-M3) {$M_3$};
    
    \coordinate (B-Offset) at (0, 0.9cm); 

    % Top Row B
    \node[compact_op] at ($(B-M2) + (B-Offset)$) (B-J1O2) {$O_{1,2}$};
    \node[compact_op, left=0.4cm of B-J1O2] (B-J1O1) {$O_{1,1}$};
    \node[compact_op, right=0.4cm of B-J1O2] (B-J1O3) {$O_{1,3}$};

    % Bottom Row B
    \node[compact_op] at ($(B-M2) - (B-Offset)$) (B-J2O2) {$O_{2,2}$};
    \node[compact_op, left=0.4cm of B-J2O2] (B-J2O1) {$O_{2,1}$};
    \node[compact_op, right=0.4cm of B-J2O2] (B-J2O3) {$O_{2,3}$};

    % Source/Sink B
    \node[compact_source, left=0.2cm of $(B-J1O1.west)!0.5!(B-J2O1.west)$] (B-S) {S};
    \node[compact_source, right=0.2cm of $(B-J1O3.east)!0.5!(B-J2O3.east)$] (B-T) {E};

    % 4. Edges B
    \draw[conj] (B-S) -- (B-J1O1); \draw[conj] (B-S) -- (B-J2O1);
    \draw[conj] (B-J1O1) -- (B-J1O2); \draw[conj] (B-J1O2) -- (B-J1O3);
    \draw[conj] (B-J2O1) -- (B-J2O2); \draw[conj] (B-J2O2) -- (B-J2O3);
    \draw[conj] (B-J1O3) -- (B-T); \draw[conj] (B-J2O3) -- (B-T);

    \draw[disj] (B-M1) -- (B-J1O1); \draw[disj] (B-M1) -- (B-J2O2);
    \draw[disj] (B-M2) -- (B-J1O2); \draw[disj] (B-M2) -- (B-J2O1);
    \draw[disj] (B-M3) -- (B-J1O3); \draw[disj] (B-M3) -- (B-J2O3);

    % ===================
    % === LABELS ===
    % ===================
    \coordinate (BottomLine) at (B-J2O2.south);
    
    \node[font=\small] at (A-J2O2 |- BottomLine) [yshift=-0.3cm] {(a) disjunctive graph};
    \node[font=\small] at (B-J2O2 |- BottomLine) [yshift=-0.3cm] {(b) proposed heterogeneous graph};

\end{tikzpicture}

    \caption{Comparison of graph representations.}
    \label{fig:jssp_combined}
    \vspace{-0.4cm}
\end{figure}
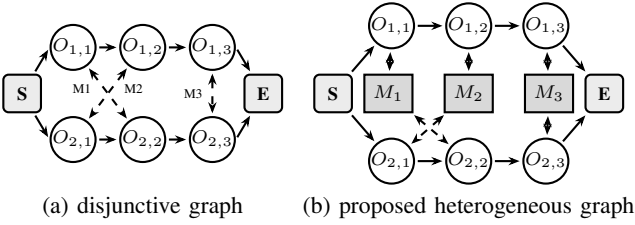

Unlike the disjunctive graph shown in Figure~\ref{fig:jssp_combined}(a), which models resource contention via dense cliques, our proposed heterogeneous graph in Figure~\ref{fig:jssp_combined}(b) utilizes a sparse structure. Our formulation is defined by two edge types: precedence edges ($\mathcal{E}_{prec}$) and assignment edges ($\mathcal{E}_{assign}$). Precedence edges are directed links that connect consecutive operations $(o_{j,k}, o_{j,k+1})$ to track job progress. Assignment edges are bidirectional links that connect each operation to its designated machine. These assignment edges establish a bipartite backbone between $\mathcal{V}_{ops}$ and $\mathcal{V}_{mch}$, enabling the propagation of contention information without the need for direct operation-to-operation conflict links.

The primary motivation for this heterogeneous design is to address the computational challenge of edge density in graph-based scheduling. Traditional disjunctive graphs form cliques among operations assigned to the same machine, leading to quadratic edge scaling $|\mathcal{E}| \sim N^2$, where $N$ denotes the total number of operations. In contrast, our approach treats machine nodes as information hubs. By decoupling direct operation-to-operation links, our structure ensures that each operation node $o \in \mathcal{V}_{ops}$ maintains a bounded degree ($deg(v) \le 3$), consisting of one assignment edge and up to two precedence edges. Consequently, the total edge count grows linearly:

\begin{equation}
|\mathcal{E}| \le (N - |\mathcal{J}|) + 2N = O(N).
\end{equation}

This linearization ensures that both the memory footprint and the computational complexity of the GNN scale linearly with problem size.

This structural efficiency provides a significant advantage in industrial online scenarios, where real-time responsiveness to dynamic events is required and the computational budget for individual decision steps is strictly limited. Our linear structural complexity limits single-step inference time to $O(N)$, a significant reduction from the $O(N^2)$ complexity of traditional approaches, maintaining computational feasibility as problem dimensions increase.

\subsection{Feature-Based Homogenization} %DONE

While our heterogeneous graph representation provides the advantage of linear scaling, it introduces the challenge of processing distinct node types with different feature spaces. To address this without the computational overhead of specialized message-passing layers (e.g., relational or heterogeneous GNNs), we employ a strategy of feature-based homogenization. This approach is motivated by recent findings by Wang et al.~\cite{wangEnablingHomogeneousGNNs2023a} that homogeneous GNN backbones possess adequate expressive power to handle heterogeneous structures if disparate node attributes are mapped into a unified latent space.

We project both operation and machine nodes into a shared latent space, augmenting their raw features with type-specific indicators. Each node $v$ is initialized with a composite feature vector $\mathbf{x}_v$ that concatenates its dynamic state attributes $\phi(v_t)$ with a learnable one-hot type embedding $\mathbf{e}_{\text{type}}$:
\begin{equation}
x_v = \mathbf{e}_{type} \parallel \phi(v_t)
\end{equation}

This allows a single homogeneous GNN backbone to project entities into orthogonal subspaces, learning distinct processing logic for operations and machines within a unified parameter space. Detailed features are provided in Table \ref{tab:node_features}.

\begin{table}[tbp]
    \vspace*{0.15cm} 
    \centering
    \caption{Unified node feature vector $\mathbf{x}_v \in \mathbb{R}^5$}
    \label{tab:node_features}
    \small % Slightly reduces font size to ensure fit
    \begin{tabularx}{\columnwidth}{@{}l X c c@{}}
        \toprule
        \textbf{idx} & \textbf{semantics} & $\bm{v \in \mathcal{V}_{ops}}$ & $\bm{v \in \mathcal{V}_{mch}}$ \\
        \midrule
        1--2 & node type & $[1, 0]^\top$ & $[0, 1]^\top$ \\
        3 & norm. lower bound & $\tilde{C}_{\text{LB}}$ & $0$ \\
        4 & sched. status & $x_{\text{mask}}$ & $0$ \\
        5 & progress ratio & $0$ & $\rho_{\text{prog}}$ \\
        \bottomrule
    \end{tabularx}
    \vspace{-0.4cm} 
\end{table}

\label{sec:architecture}

By mapping all entities into this shared feature space, we treat the shop floor as a formally homogeneous graph \mbox{$\mathcal{G} = (\mathcal{V}, \mathcal{E})$}, where node-type distinctions are handled implicitly via the learnable embeddings. On this representation, we employ a $K$-layer GIN \cite{xuHowPowerfulAre2019}. Node embeddings are initialized as $\mathbf{h}_v^{(0)} = \mathbf{x}_v$ and updated via a multi-layer perceptron $\text{MLP}^{(k)}$:
\begin{equation}
    h_v^{(k)} = \text{MLP}^{(k)} \left( (1 + \epsilon^{(k)}) \cdot h_v^{(k-1)} + \sum_{u \in \mathcal{N}(v)} h_u^{(k-1)} \right)
\end{equation}
where $\epsilon^{(k)}$ is a learnable scalar and $\mathcal{N}(v)$ denotes the set of neighbors within the sparse unified graph structure.

By stacking $K$ layers, the receptive field of each node expands to its $K$-hop neighborhood. For example, at $K=2$, an operation node $O_{i,j}$ integrates information not only from its direct machine but also from all competing operations assigned to the same machine. 
% This architecture enables multi-channel reasoning across three critical structural dimensions:
This architecture supports three forms of reasoning: precedence edges propagate temporal information along jobs, operation-machine aggregation captures resource contention, and bidirectional machine-operation feedback exposes queue density to waiting operations.
% \begin{itemize}
%     \item Intra-job precedence flow ($O_{j,k} \rightarrow O_{j,k+1}$) propagates temporal information downstream through directed edges, allowing the network to update the features of all subsequent operations whenever a delay or completion occurs on a job's critical path.
%     \item Resource contention aggregation ($O \rightarrow \mathcal{M}$) utilizes machine nodes as information hubs that collect feature vectors from all assigned operations to quantify global resource demand and estimate potential shop floor congestion.
%     \item Resource-state feedback ($\mathcal{M} \rightarrow O$) completes the communication loop via bidirectional edges, ensuring that machine status and queue density are reflected back to waiting operations so the policy can account for real-time contention from competing jobs.
% \end{itemize}

To support both local decision-making and global state evaluation, the architecture produces two distinct levels of representation. We compute a global graph embedding $h_G$ via sum-pooling over all final node embeddings:
\begin{equation}
    h_G = \text{READOUT}\left( \{ h_v^{(K)} \mid v \in \mathcal{V} \} \right) = \sum_{v \in \mathcal{V}} h_v^{(K)}
\end{equation}

Unlike mean-pooling, sum-pooling preserves information about the total graph size. In the context of scheduling, this magnitude correlates with the number of pending operations and the remaining workload, providing a necessary global scale of the shop floor state. The framework yields a set of node-level embeddings $\{h_v^{(K)}\}$ for local action selection and a single graph-level embedding $h_G$ for state-value estimation.

\subsection{Policy Optimization}
\label{sec:policy_architecture}

We employ an actor-critic framework trained via PPO, where the GIN backbone parameterizes both the policy $\pi_\theta(a_t|s_t)$ and the critic $V_\phi(s_t)$. The network bifurcates into two specialized heads that process the learned node embeddings $h_v^{(K)}$ and the global graph embedding $h_G$ to capture multi-scale shop floor dynamics.

The critic (Value Head) estimates the expected return $V(s_t)$, representing the expected sum of discounted future rewards when starting from state $s_t$. This estimation is derived purely from the global graph state, providing a workload-aware baseline for advantage estimation:

\begin{equation}
    V(s_t) = \text{MLP}_{\text{critic}}(h_G)
\end{equation}

The actor (policy head) computes action probabilities over the operation nodes $\mathcal{V}_{\text{ops}}$. To capture both local scheduling constraints and global bottlenecks, we concatenate each node's local embedding with the broadcasted global context:
\begin{equation}
    z_v = h_v^{(K)} \parallel h_G, \quad \forall v \in \mathcal{V}_{\text{ops}}
\end{equation}
By concatenating $h_v^{(K)}$ with $h_G$, we provide the actor with a relational anchor, allowing it to evaluate the importance of a local task relative to the total remaining work. These context-aware vectors are projected to logits $l_v = \text{MLP}{\text{actor}}(z_v)$. To enforce precedence constraints, we apply a dynamic action mask $\mu_t$, where $\mu_{t,v} = 0$ for feasible operations and $-\infty$ otherwise. The final policy is computed as a masked softmax:
\begin{equation}
    \pi_\theta(a_t \mid s_t) = \text{softmax}(\mathbf{l} + \boldsymbol{\mu}_t)
\end{equation}

\begin{algorithm}[b]
\caption{PPO training with heterogeneous graph embedding.}
\label{alg:training}
\small
\begin{algorithmic}[1]
\State \textbf{Input:} distribution $\mathcal{D}_{\text{inst}}$ of $|\mathcal{J}| \times |\mathcal{M}|$ instances
\State \textbf{Init:} actor $\pi_\theta$, critic $V_\phi$, buffer $\mathcal{B}$
\For{episode $k = 1 \dots K$}
    \State sample instance $\mathcal{I} \sim \mathcal{D}_{\text{inst}}$; \ $s_0 \leftarrow \text{RESET}(\mathcal{I})$
    \State $t \leftarrow 0$
    \While{$s_t$ is not terminal}
        \State $G_t \leftarrow \textsc{ConstructGraph}(s_t)$
        \State $\mathcal{M}_t \leftarrow \textsc{GetActionMask}(s_t)$
        
        \State $h_v, h_G \leftarrow \text{GIN}_\theta(G_t)$ \Comment{node \& global embed.}
        
        \State $v_t \leftarrow \text{MLP}_\phi(h_G)$ \Comment{critic value estimate}
        
        \State $z_v \leftarrow [h_v \parallel h_G]$ \Comment{actor context}
        \State $\pi_t \leftarrow \text{softmax}(\text{MLP}_\theta(z_v) + \log \mathcal{M}_t)$
        
        \State sample action $a_t \sim \pi_t$
        \State $s_{t+1}, r_t \leftarrow \textsc{step}(s_t, a_t)$
        
        \State store $(s_t, a_t, r_t, v_t, \log \pi_t(a_t))$ in $\mathcal{B}$
        \State $t \leftarrow t + 1$
    \EndWhile
    \State compute GAE advantages $\hat{A}_t$ using stored $v_t$
    \State update $\theta, \phi$ via PPO clip loss on $\mathcal{B}$
\EndFor
\end{algorithmic}
\end{algorithm}
% \vspace{-1.1em} 

The procedural execution is formalized in Algorithm \ref{alg:training}. Parameters are optimized to maximize the PPO clipped objective:

\begin{equation}
\begin{split}
    L^{CLIP}(\theta) = \hat{\mathbb{E}}_t \Big[ \min \Big( & r_t(\theta)\hat{A}_t, \\ 
    & \text{clip}(r_t(\theta), 1-\epsilon, 1+\epsilon)\hat{A}_t \Big) \Big]
\end{split}
\end{equation}
where $r_t(\theta)$ is the probability ratio between the new and old policies. The term $\hat{A}_t = \sum_{k=0}^{\infty} (\gamma \lambda)^k \delta_{t+k}$ denotes the Generalized Advantage Estimation (GAE), which relies on the temporal difference error $\delta_t = r_t + \gamma V(s_{t+1}) - V(s_t)$. This objective ensures stable policy updates by restricting the ratio within a small interval $[1-\epsilon, 1+\epsilon]$, leveraging the workload-sensitive baseline from the critic to reduce variance across diverse JSSP instances.

\begin{figure*}
\vspace*{0.05cm} 

\centering
\begin{tikzpicture}[
    % --- Inherited Styles from JSSP Graph ---
    node distance=0.6cm and 0.6cm,
    % Operations (Circles)
    compact_op/.style={ 
        circle, draw=black, thick, fill=white,
        minimum size=0.45cm, 
        inner sep=0pt, font=\scriptsize\bfseries 
    },
    % Machines/Static Elements (Gray Rects)
    compact_machine/.style={ 
        rectangle, draw=black, thick, fill=gray!30, 
        minimum size=0.5cm, 
        font=\scriptsize\bfseries
    },
    % Processing Blocks (GIN layers - styled like Source/Sink but white for logic)
    compact_block/.style={ 
        rectangle, draw=black, thick, fill=white, 
        rounded corners=2pt, 
        minimum width=0.8cm, minimum height=0.8cm,
        font=\scriptsize\bfseries, align=center
    },
    % Vectors (Embeddings)
    compact_vec/.style={
        rectangle, draw=black, thick, fill=white,
        minimum width=0.8cm, minimum height=0.35cm,
        inner sep=0pt
    },
    % MLP Heads (Trapeziums)
    compact_mlp/.style={
        trapezium, trapezium angle=80,
        draw=black, thick, fill=white,
        minimum height=0.5cm, minimum width=1.0cm,
        align=center, rotate=-90,
        font=\scriptsize\bfseries
    },
    % Edges
    conj/.style={ 
        ->, >={Stealth[length=1.5mm]}, thick, black,
        shorten >=1pt, shorten <=1pt 
    },
    disj/.style={ 
        <->, >={Stealth[length=1.5mm]}, dashed, thick, black,
        shorten >=1pt, shorten <=1pt 
    },
    aux/.style={
        ->, >={Stealth[length=1.5mm]}, dashed, thick, black!60,
        shorten >=1pt, shorten <=1pt
    },
    % Text Headers
    header_text/.style={
        font=\bfseries\small,
        align=center,
        yshift=-1.8cm
    },
    route/.style={thick, black},
    conj_tight/.style={
    ->, >={Stealth[length=1.5mm]}, thick, black,
    shorten >=0pt, shorten <=0pt
    }
]

    % =========================================================================
    % 1. GRAPH STATE
    % =========================================================================
    \begin{scope}[local bounding box=scope1]
        % Operations (Vertical)
        \node[compact_op] (O1) at (0, 0.8) {$O_1$};
        \node[compact_op] (O2) at (0, 0) {$O_2$};
        \node[compact_op] (O3) at (0, -0.8) {$O_n$};
        
        % Machines (Offset)
        \node[compact_machine] (M1) at (1.0, 0.4) {$M_1$}; 
        \node[compact_machine] (M2) at (1.0, -0.4) {$M_2$}; 
        
        % Edges
        \draw[conj] (O1) -- (O2);
        \draw[conj] (O2) -- (O3);
        
        \draw[disj] (O1) -- (M1);
        \draw[disj] (O2) -- (M2);
        \draw[disj] (O3) -- (M1);
    \end{scope}

    % =========================================================================
    % 2. BACKBONE
    % =========================================================================
    \begin{scope}[shift={(3.1,0)}, local bounding box=scope2]
    \node[compact_block] (L1) at (-0.9, 0) {GIN\\1};
    \node[font=\bfseries\scriptsize, inner sep=1pt] (dots) at (0.1, 0) {$\cdots$};
    \node[compact_block] (LK) at (0.9, 0) {GIN\\$K$};
    
    \draw[conj] (L1) -- (dots);
    \draw[conj] (dots) -- (LK);
    
    \end{scope}

    % =========================================================================
    % 3. EMBED
    % =========================================================================
    \begin{scope}[shift={(5.6,0)}, local bounding box=scope3]
    \node[compact_op] (src_op) at (-0.6, 0.5) {$O$};
    \node[compact_machine] (src_mch) at (-0.6, -0.5) {$M$};
    
    \node[compact_vec] (vec1) at (0.6, 0.5) {}; 
    \fill[gray!30] (vec1.north west) rectangle ($(vec1.south west)+(0.25,0)$);
    \draw[thick] (vec1.north west) rectangle (vec1.south east);
    \node[right, font=\tiny] at (vec1.east) {$\phi(o)$};
    
    \node[compact_vec] (vec2) at (0.6, -0.5) {}; 
    \fill[gray!30] ($(vec2.north east)-(0.25,0)$) rectangle (vec2.south east); 
    \draw[thick] (vec2.north west) rectangle (vec2.south east);
    \node[right, font=\tiny] at (vec2.east) {$\phi(m)$};
    
    \draw[conj] (src_op) -- (vec1);
    \draw[conj] (src_mch) -- (vec2);
    
    \draw[decorate, decoration={brace, amplitude=4pt}, thick, black] 
        (1.7, 0.8) -- (1.7, -0.8);
    
    \end{scope}

    % =========================================================================
    % 4. PPO HEADS
    % =========================================================================
\begin{scope}[shift={(9.2,0)}, local bounding box=scope4]
    \coordinate (s4_split) at (-1.2, 0); 
    
    \draw[conj] (scope3.east |- 0,0) -- (s4_split);

    % --- Top Branch (actor) ---
    \node[circle, draw=black, thick, inner sep=1pt, minimum size=0.4cm, font=\scriptsize\bfseries] (cat) at (-0.2, 0.8) {$\|$}; 
    \node[compact_mlp] (actor) at (0.6, 0.8) {act}; 
    
    % \node[below=0.4cm of actor, xshift=0.3cm, font=\tiny] {local};

    % Mask
    \node[rectangle, draw=black, thick, pattern=north east lines, pattern color=black!60, minimum width=0.15cm, minimum height=0.5cm] (mask) at (1.4, 0.8) {}; 
    \node[above, font=\tiny] at (mask.north) {msk};
    \node[right=0.05cm of mask, font=\bfseries\scriptsize] {$\pi$};

    \draw[conj] (s4_split) |- (cat.west);
    \draw[conj] (cat) -- (actor);
    \draw[conj] (actor) -- (mask);

    % --- Bottom Branch (critic) ---
    \node[compact_op] (pool) at (-0.2, -0.8) {$\Sigma$}; 
    \node[below=0.05cm of pool, font=\tiny, color=black] {global};

    \node[compact_mlp] (critic) at (0.6, -0.8) {crit}; 
    \node[compact_op] (val) at (1.4, -0.8) {$V$}; 

    \draw[conj] (s4_split) |- (pool.west);
    \draw[conj] (pool) -- (critic);
    \draw[conj] (critic) -- (val);

    % --- Routing (Global info to actor) ---
    \draw[aux] (pool.north) -- (cat.south);
    \node[right, font=\tiny] at (-0.2, 0.0) {$h_G$};
\end{scope}

    % =========================================================================
    % GLOBAL ARROWS & LABELS
    % =========================================================================
    \draw[conj] (scope1.east |- 0,0) -- (scope2.west |- 0,0);
    \draw[conj] (scope2.east |- 0,0) -- (scope3.west |- 0,0);
    
    % Labels
    \node[header_text] at (scope1 |- 0, -0.1) {1. Graph State};
    \node[header_text] at (scope2 |- 0, -0.1) {2. Backbone};
    \node[header_text] at (scope3 |- 0, -0.1) {3. Embed};
    \node[header_text] at (scope4 |- 0, -0.1) {4. PPO Heads};

\end{tikzpicture}
    \caption{Architecture of the policy optimization network. We utilize a heterogeneous graph embedding followed by a homogeneous GIN backbone. The actor and critic heads share features but use distinct masking and pooling mechanisms.}
    \label{fig:policy_architecture}
    \vspace{-0.5cm}
\end{figure*}
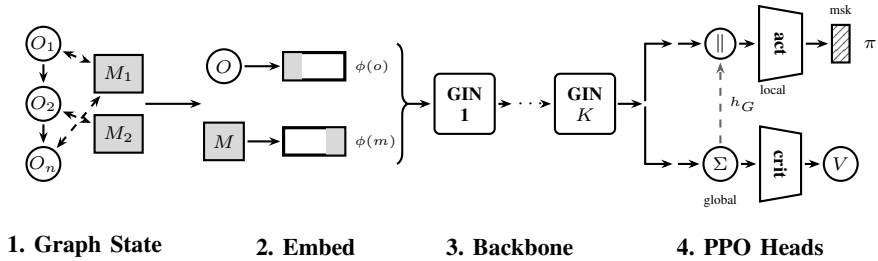

Figure~\ref{fig:policy_architecture} summarizes the resulting data flow from graph construction and feature-based homogenization through the homogeneous GIN backbone to the masked actor and pooled critic heads.

\section{Experiments and Results} \label{sec:experiments}
\subsection{Experimental Setup} \label{sec:setup}  
We design our experimental evaluation to validate the general efficacy of our proposed unified graph framework and systematically investigate the zero-shot generalization capabilities of the learned policies.

To investigate how the training distribution affects generalization, we train multiple independent agents across a range of problem geometries $\mathcal{J} \times \mathcal{M}$. Evaluation is conducted on a comprehensive testbed of 1,820 instances covering diverse scales and configurations. For architectural baseline validation, we utilize standard historical datasets, namely the Taillard \cite{taillardBenchmarksBasicScheduling1993} and Lawrence \cite{lawrence1984resource} benchmarks. We additionally introduce a custom synthetic suite to assess topological sensitivity, comprising 100 instances per configuration and spanning square geometries ($5 \times 5$ to $20 \times 20$) as well as rectangular layouts ($10 \times 5$ up to $100 \times 20$) with systematically varied job-to-machine ratios. Synthetic instances follow the standard random JSSP protocol: machine routes are sampled as random permutations, processing times are drawn independently from $\mathcal{U}\{1,99\}$, and job-to-machine ratios are varied by changing the number of jobs for fixed machine sets. All generator settings and seeds are included in the released code repository.

\subsection{Implementation Details}
\label{sec:implementation}
Our framework is implemented using PyTorch Geometric \cite{feyFastGraphRepresentation2019} and TorchRL \cite{bouTorchRLDatadrivenDecisionmaking2023}. To facilitate further research and reproducibility, we published the source code, training environments, and synthetic datasets at \url{https://github.com/proto-lab-ro/unified-graph-rl-jssp}. 

We utilize a $K=3$ layer GIN backbone with node features normalized by the estimated instance horizon to ensure numerical stability. The depth $K$ controls the receptive field of each operation node: one message-passing step captures direct precedence and machine-assignment information, while two steps expose competing operations through operation-machine-operation paths. Larger $K$ propagates information over longer precedence and contention neighborhoods, but also increases inference cost and may reduce embedding distinctiveness due to over-smoothing. We therefore choose $K=3$ as a trade-off to capture immediate precedence relations, machine contention, and short-range downstream effects while preserving low-latency inference. To maintain consistent gradient variance across varying problem scales, we employ a dynamic rollout batch size proportional to the total number of operations. Training is performed using the Adam optimizer with a linear decaying learning rate. Comprehensive hyperparameters and architectural dimensions are provided in Table \ref{tab:hyperparameters}.

\begin{table}[h]
\centering
\caption{Hyperparameter configuration}
\label{tab:hyperparameters}
\small
\begin{tabular}{l l}
\toprule
\textbf{Parameter} & \textbf{Value} \\
\midrule
\multicolumn{2}{l}{\textit{Optimization (PPO)}} \\
Learning Rate & $3 \times 10^{-4}$ (Linear Decay) \\
Batch Size & $4 \times (|\mathcal{J}| \times |\mathcal{M}|)$ transitions \\
Discount / GAE & $\gamma=1.0, \ \lambda=1.0$ \\
Coefficients & Clip $\epsilon=0.2, \ c_{ent}=0.01$ \\
\midrule
\multicolumn{2}{l}{\textit{Architecture}} \\
Backbone & GIN ($K=3, \ d_{hidden}=64$) \\
Policy / Value Head & MLP ($64 \to 64 \to 1$, Tanh) \\
\bottomrule
\end{tabular}
\end{table}

\subsection{Comparative Performance Analysis}
\label{sec:results_standard}

To establish a baseline, we benchmark against established PDRs, including Most Work Remaining (MWKR), Most Operations Remaining (MOR), and Flow Due Date / Most Work Remaining (FDD/MWR), as well as the state-of-the-art RL framework by Zhang et al. \cite{zhangLearningDispatchJob2020}. Other recent works, such as Hameed and Schwung \cite{hameedGraphNeuralNetworksbased2023} and Tassel et al. \cite{tasselReinforcementLearningEnvironment2021}, are excluded due to incompatible decentralized or instance-specific training protocols.

Table \ref{tab:standard_benchmarks} summarizes results on the Taillard benchmarks and our synthetic validation set. Our composite configuration, which selects the best size-specific agent per geometry, denoted as \textbf{Ours (comp)}, consistently outperforms all PDRs and also surpasses Zhang et al. Notably, we observe that even a model trained on only a single geometry ($20 \times 20$), denoted as \textbf{Ours (single)}, already outperforms the PDRs and achieves parity with Zhang et al. In the few cases where Zhang et al. reports lower makespans, the difference may reflect their more flexible construction protocol, which permits insertion into partial schedules, whereas our framework follows an online-style dispatching setting with irrevocable decisions. Overall, the results show state-of-the-art performance while preserving the scalability advantages of the proposed formulation.

\begin{table*}[tbp]
\vspace*{0.1cm} % <--- Forces the entire table down
    \centering
\caption{Average makespan on synthetic and standard benchmarks. Mean $\pm$ std.; best non-reference values in \textbf{bold}.}    \label{tab:standard_benchmarks}
    \small
    \begin{tabularx}{\textwidth}{l c c c c c c}
        \toprule
        \textbf{Instance Size} & \textbf{MWKR} & \textbf{MOR} & \textbf{FDD$/$MWR} & \textbf{Zhang et al. \cite{zhangLearningDispatchJob2020}} & \textbf{Ours (comp)} & \textbf{Ours
        (single)} \\
        \midrule
        \multicolumn{7}{l}{\textit{Synthetic Validation Set with 100 Instances per Size}} \\
        $6\times6$ & $653.8 \pm 80.7$ & $611.7 \pm 70.4$ & $591.5 \pm 62.8$ & $574.09$ & $\textbf{533.5} \pm \textbf{50.3}$ & $585.1 \pm 66.3$ \\
        $10\times10$ & $1192.6 \pm 120.9$ & $1086.5 \pm 103.0$ & $1031.3 \pm 85.5$ & $988.58$ & $\textbf{949.2} \pm \textbf{75.8}$ & $1008.0 \pm 90.8$ \\
        $15\times15$ & $1866.9 \pm 150.4$ & $1691.1 \pm 116.7$ & $1563.3 \pm 101.7$ & $1504.79$ & $\textbf{1464.0} \pm \textbf{85.1}$ & $1516.5 \pm 98.1$ \\
        $20\times20$ & $2555.0 \pm 169.0$ & $2298.4 \pm 147.0$ & $2093.3 \pm 111.7$ & $2007.76$ & $\textbf{1987.8} \pm \textbf{90.3}$ & $2079.7 \pm 115.7$ \\
        $30\times20$ & $3286.6 \pm 200.8$ & $2918.5 \pm 175.4$ & $2684.1 \pm 124.1$ & $\textbf{2508.27}$ & $2512.4 \pm 107.3$ & $2652.2 \pm 143.7$ \\
        \midrule
        \multicolumn{7}{l}{\textit{Taillard Benchmark \cite{taillardBenchmarksBasicScheduling1993}}} \\
        $15\times15$ & $1896.1 \pm 107.8$ & $1735.7 \pm 88.2$ & $1577.4 \pm 56.8$ & $1547.4$ & $\textbf{1488.9} \pm \textbf{21.6}$ & $1538.3 \pm 39.7$ \\
        $20\times15$ & $2178.9 \pm 116.6$ & $1998.5 \pm 103.4$ & $1836.7 \pm 96.8$ & $1774.7$ & $\textbf{1755.4} \pm \textbf{70.8}$ & $1845.2 \pm 93.4$ \\
        $20\times20$ & $2633.9 \pm 163.0$ & $2334.8 \pm 102.5$ & $2164.7 \pm 81.1$ & $2128.1$ & $\textbf{2053.8} \pm \textbf{41.7}$ & $2156.5 \pm 66.0$ \\
        $30\times15$ & $2985.9 \pm 216.5$ & $2567.6 \pm 165.1$ & $2511.2 \pm 151.7$ & $2378.8$ & $\textbf{2289.1} \pm \textbf{82.8}$ & $2346.5 \pm 120.8$ \\
        $30\times20$ & $3265.7 \pm 139.9$ & $3010.4 \pm 139.5$ & $2726.4 \pm 100.8$ & $2603.9$ & $\textbf{2594.2} \pm \textbf{92.1}$ & $2756.0 \pm 119.8$ \\     
        $50\times15$ & $4444.9 \pm 290.8$ & $3714.1 \pm 195.3$ & $3723.7 \pm 169.8$ & $3393.8$ & $\textbf{3349.3} \pm \textbf{136.3}$ & $3423.3 \pm 159.7$ \\ 
        $50\times20$ & $4677.4 \pm 210.5$ & $4143.6 \pm 232.6$ & $3845.5 \pm 177.0$ & $3593.9$ & $\textbf{3566.9} \pm \textbf{137.7}$ & $3730.2 \pm 212.9$ \\
        $100\times20$ & $8040.3 \pm 235.4$ & $6728.8 \pm 307.8$ & $6837.4 \pm 262.1$ & $\textbf{6097.6}$ & $6263.4 \pm 274.8$ & $6423.0 \pm 261.4$ \\
        \bottomrule
    \end{tabularx}
    \par\medskip
    \footnotesize \textit{Note:} Zhang et al. and Ours (comp) employ shape-specific models tailored to the instance geometry. Ours (single) is a single generalist policy trained exclusively on $20 \times 20$ to evaluate zero-shot generalization across all scales.
    \vspace{-0.5cm} % <--- Adjust this value to pull the text up
\end{table*}

\subsection{Scalability and Zero-Shot Transfer}
\label{sec:generalization}

Beyond predictive performance, practical deployment in industrial environments requires scalable computational effort. Owing to its unified graph representation, our approach inherently achieves linear scaling with problem size, making it computationally tractable as instance dimensions grow.
Furthermore, in dynamic real-world settings, the ability to operate without retraining when instance dimensions change provides an additional practical advantage. We therefore evaluate zero-shot transfer as a measure of cross-scale adaptability.

For this analysis, we benchmark our most robust generalist model, Ours (single), against PDRs, which remain the industry standard due to their size-agnostic and deployment-ready nature. We do not include Zhang et al. \cite{zhangLearningDispatchJob2020} in this setting, as their approach relies on geometry-specific training and was not designed for fully online, scale-agnostic application.

\begin{figure}[tbp]
    \centering
    \begin{subfigure}{0.48\textwidth}
        \includegraphics[width=\linewidth]{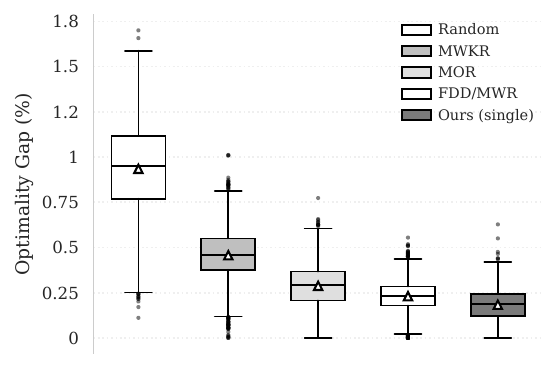}
        \caption{Performance across the test suite of 1,820 instances}
        \label{fig:optimality_gap}
    \end{subfigure}
    \\
    \begin{subfigure}{0.48\textwidth}
        \includegraphics[width=\linewidth]{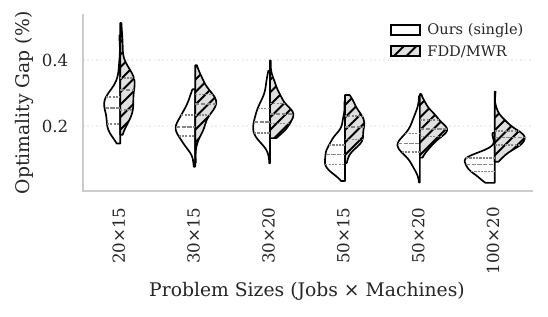}
        \caption{Zero-shot scalability study on 100 instances per size}
        \label{fig:scalability}
    \end{subfigure}

    \caption{Size-agnostic performance comparison.}
    
    \label{fig:combined_results}
    \vspace{-0.5cm}
\end{figure}

The primary metric is the optimality gap, measuring deviation from the best-known lower bound. As shown in Figure \ref{fig:optimality_gap}, Ours (single) consistently outperforms standard PDRs such as MWKR and MOR. While these simple rules exhibit high variance, the more sophisticated FDD/MWR rule \cite{selsComparisonPriorityRules2012a} provides a stable baseline that matches our model on smaller instances.

However, Figure \ref{fig:scalability} reveals a divergence in scalability. As problem sizes increase toward $100 \times 20$, our policy achieves a notably lower median optimality gap and reduced variance compared to FDD/MWR. This confirms that the relative advantage of our framework over PDRs grows with the problem scale. While fixed-priority heuristics rely on local decision rules, our unified graph architecture disentangles scheduling logic from global problem size, allowing the agent to leverage a global view of operation interdependencies. This advantage is particularly relevant for industrial applications, as the performance edge of our approach becomes most pronounced on the large-scale instances that characterize complex real-world production environments.

\subsection{Hypothesis of Structural Saturation}
\label{sec:saturation_hypothesis}
\begin{figure}[tbp]
    \centering
    \begin{subfigure}[b]{0.49\textwidth}
        \centering
        \includegraphics[width=\linewidth]{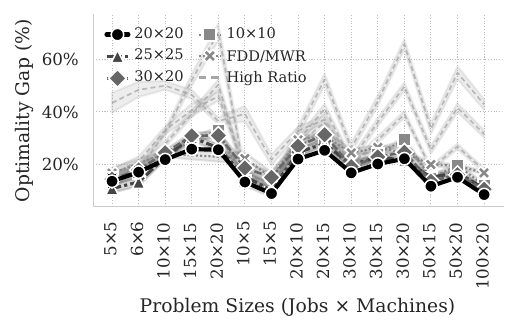}
        \caption{Performance by instance size and shape}
        \label{fig:sub_sorted_size}
    \end{subfigure}
    \hfill 
    \begin{subfigure}[b]{0.49\textwidth}
        \centering
        \includegraphics[width=\linewidth]{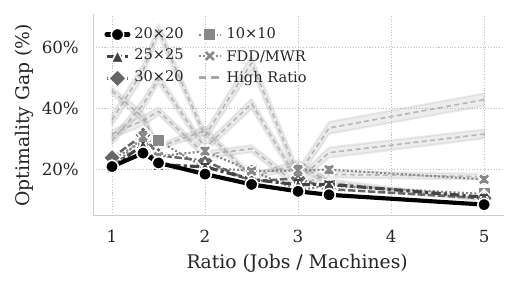}
        \caption{Performance by job-to-machine ratio}
        \label{fig:sub_ratio}
    \end{subfigure}
    \caption{Impact of Training Topology on Generalization.}
    \label{fig:generalization_analysis}
    \vspace{-0.6cm}
\end{figure}

Through the evaluation of agents trained on various geometries across test scales from $5 \times 5$ to $100 \times 20$, the $20 \times 20$ configuration emerged as the most resilient generalist among the evaluated training geometries. As shown in Figure \ref{fig:sub_sorted_size}, this agent consistently achieves the lowest average optimality gap across all instance shapes. Furthermore, Figure \ref{fig:sub_ratio} confirms that its performance dominates across a vast \mbox{job-to-machine} ratio spectrum, outperforming agents trained on both smaller and larger geometries.

As an explanation, we propose a \textit{hypothesis of structural saturation}. This hypothesis is motivated by the easy-hard-easy phase transition in combinatorial optimization, where difficulty is driven by constraint density rather than absolute size. We argue that policy generalization is determined by the training instance topology rather than the absolute problem scale. We categorize this topology into three distinct regimes based on the job-to-machine ratio ($\mathcal{J}/\mathcal{M}$):

In the undersaturated regime ($\mathcal{J}/ \mathcal{M} \ll 1$), sparse conflicts between jobs allow simple, myopic rules to succeed. Because agents trained in this setting face minimal penalties for sub-optimal decisions, they fail to internalize the robust conflict-resolution strategies or long-horizon logic required for more complex environments.

As the system transitions into the saturated regime ($\mathcal{J}/ \mathcal{M} \approx 1$), it reaches the point of peak combinatorial hardness. As illustrated in Figure \ref{fig:sub_ratio}, the optimality gap is generally higher here, indicating that these instances are the most challenging. This intense resource contention compels the agent to learn invariant scheduling principles that remain effective even when the problem scale increases.

In contrast, the supersaturated regime ($\mathcal{J}/ \mathcal{M} \gg 1$) behaves increasingly like a stable queueing network. While the problem dimensions are large, the combinatorial difficulty actually decreases because high machine utilization becomes a sufficient condition for near-optimality. Agents trained exclusively in this regime may overfit to these flow-dominated dynamics, resulting in policies that lack the complex conflict-resolution logic needed to navigate the tighter constraints of quadratic benchmarks.

Our results identify the $20 \times 20$ configuration as a structural saturation point within our benchmark suite. This implies that the agent treats massive rectangular instances, such as $100 \times 20$, as a sequential concatenation of saturated sub-problems. In an industrial context, this suggests that large-scale scheduling is best addressed by training at the saturation point rather than the target scale. This approach ensures policy robustness against high-contention scenarios while drastically reducing computational overhead by avoiding training on massive, yet combinatorically simpler, topologies.

\section{Conclusion}
We proposed a unified graph-based RL framework for the JSSP that achieves high-quality scheduling policies with linear computational complexity. By employing feature-based homogenization, our approach eliminates the overhead of heterogeneous layers while maintaining strong relational inductive biases. Experimental results demonstrate that our agent significantly outperforms traditional PDRs and achieves state-of-the-art performance on public benchmarks.

Crucially, we identified a structural saturation point at the $20 \times 20$ configuration, where the internal constraint density allows the agent to learn scale-invariant scheduling logic. This enables robust zero-shot generalization to industrial-scale instances such as $100 \times 20$, effectively treating them as a concatenation of saturated sub-problems.

By combining computational feasibility, low-latency online inference, and zero-shot generalization, our framework satisfies the primary requirements for deployment in real-world industrial environments. Future work will extend this approach to stochastic settings with machine breakdowns and variable job arrival rates to further bridge the gap between theoretical optimization and dynamic industrial application.

\vspace*{-5px} % <--- Add this BEFORE the environment to push it down. Adjust value as needed!

\addtolength{\textheight}{-0cm}   % This command serves to balance the column lengths
                                  % on the last page of the document manually. It shortens
                                  % the textheight of the last page by a suitable amount.
                                  % This command does not take effect until the next page
                                  % so it should come on the page before the last. Make
                                  % sure that you do not shorten the textheight too much.

\bibliographystyle{IEEEtran}

\bibliography{references}

\end{document}